\documentclass{article}
\usepackage{microtype}
\usepackage{graphicx}
\usepackage{booktabs}
\usepackage{xcolor}
\usepackage{hyperref}

\usepackage[preprint]{icml2026}
\usepackage{amsmath}
\usepackage{amssymb}
\usepackage{float}
\usepackage[capitalize,noabbrev]{cleveref}

\newcommand{\KL}{\mathrm{KL}}

\newcommand{\Sref}[1]{Appendix~\ref{#1}}

\icmltitlerunning{Which Signal Should Select a Compressed LLM?}

\begin{document}

\twocolumn[
  \icmltitle{Which Signal Should Select a Compressed LLM? An Empirical Study of\\
    Pre- and Post-Recovery Selection in Block-Replacement Search}
  \begin{icmlauthorlist}
    \icmlauthor{Yuefeng Xu}{gs}
    \icmlauthor{Jun Yu}{ist}
  \end{icmlauthorlist}
  \icmlaffiliation{gs}{Graduate School of Science and Technology, Niigata University, Niigata, Japan}
  \icmlaffiliation{ist}{Institute of Science and Technology, Niigata University, Niigata, Japan}
  \icmlcorrespondingauthor{Jun Yu}{Yujun@ie.niigata-u.ac.jp}
  \icmlkeywords{Model compression, large language models, architecture search, knowledge distillation}
  \vskip 0.3in
]
\printAffiliationsAndNotice{}

\begin{abstract}
Block-replacement compression of large language models distils cheaper replacements for every layer, searches for
the combination that minimises a distillation objective measured before recovery, and then recovers the chosen
model by fine-tuning. Within one pool of block-replacement candidates and a fixed recovery budget, we compare
several selection signals, measured before and after a short recovery, with the candidates' downstream accuracy after
full recovery. On six settings from two model families (Qwen2.5-0.5B, 1.5B and 3B; Llama-3.2-1B and 3B; full fine-tuning and LoRA),
with seven searched structures and two whole-layer-removal baselines at three latency budgets, and with every signal
scored against the final accuracy on the same held-out task examples, the mean calibration KL that the search
optimises orders the strong candidates significantly better than chance in none of the settings at our main
threshold (in one at either alternative threshold), and below chance on Qwen2.5-3B. After 100 of the 2{,}000 recovery steps, zero-shot accuracy on held-out examples is the only signal that
orders the candidates significantly better than chance in all six settings (80--94\% of the decided pairs, also when
selecting on four tasks and evaluating on a fifth), while the divergence and the training loss used by existing
training-aware selection are significant in two or three; pooled over the settings it orders 13--19 points more pairs
correctly than each of them and 30 more than the search objective. Used as a selection rule, its regret (averaged over the two example halves) stays within
0.81 points of the best searched structure in all 18 setting--budget cells, whereas always deploying the best single
search method falls up to 1.70 points behind and picking by the search objective up to 2.85, because the best method
changes from cell to cell; zero-shot accuracy before recovery, which costs no training, comes second (mean 0.39 against
0.27 points). In prospective runs on three models, where the rules select from 16 near-optimal
structures per budget and only the picks are fully recovered, this advantage does not carry over: the rule is
significantly worse than a searched structure in three of nine cells and better than all of them in one, where it picks
the pool's surrogate optimum, while the zero-step rule, run on two of the models, is never significantly worse. Stacking
one block type compounds the divergence on both Qwen models but the task-accuracy drop only on the 1.5B model. An
exploratory statistic, the concentration of the pre-recovery divergence on few tokens, replicated on Llama-3.2-1B but
not on Llama-3.2-3B. A surrogate corrected for compounding lowers the pre-recovery objective without improving the
recovered model.
\end{abstract}

\section{Introduction}
\label{sec:intro}
Structured compression makes a pretrained language model cheaper by removing or replacing parts of it and then
recovering the lost quality with a short fine-tuning~\citep{ma2023llmpruner,xia2024sheared,muralidharan2024minitron}.
Block-replacement methods such as Puzzle~\citep{bercovich2024puzzle}, following blockwise distillation for vision
models~\citep{li2020dna,moons2021donna,molchanov2022lana}, give every layer cheaper alternatives (fewer attention
heads, a pruned or low-rank MLP, no attention, no MLP, no layer), train each to mimic the original layer, and select
one alternative per layer under a latency budget by optimising a distillation objective of the assembled model
measured \emph{before} recovery. Prior work~\citep{xu2026additivity} showed that the additive surrogates used for
this optimisation fail at strong compression and repaired them.

\begin{figure}[t]
\centering
\includegraphics[width=\columnwidth]{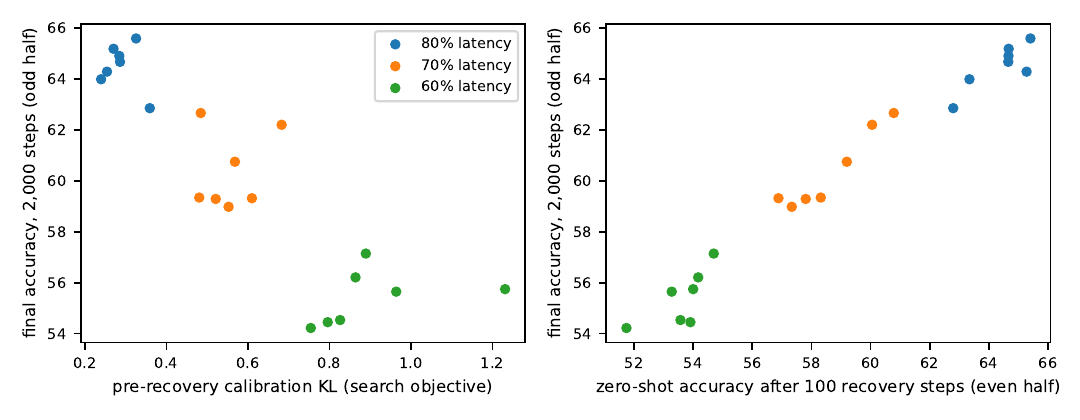}
\caption{Qwen2.5-3B, seven searched structures per latency budget. Left: the pre-recovery objective that searches
minimise does not track the structures' accuracy after recovery. Right: zero-shot accuracy after 100 of the
2{,}000 recovery steps (even-indexed task examples) orders them by final accuracy (odd-indexed examples).}
\label{fig:teaser}
\end{figure}

We ask a different question: how well does the pre-recovery objective, or any other signal, rank complete
candidates the way recovery does? The deployed model is the recovered one. Several methods already look beyond the
pre-recovery objective. Minitron~\citep{muralidharan2024minitron} and DarwinLM~\citep{tang2026darwinlm} retrain
candidates briefly and rank them by a validation loss or by the KL divergence to the parent; DarwinLM shows
qualitatively, for 16 candidates of one model, that the KL after short training predicts the KL after longer training.
EvoPress~\citep{sieberling2025evopress} evaluates candidates without training, by the KL on growing calibration samples.
Puzzle~\citep{bercovich2024puzzle} compares LM loss, KL and downstream accuracy as \emph{block} scores and finds KL better
than loss. Studies of compressed models warn that perplexity can hide lost
ability~\citep{jaiswal2024llmkick,chen2024streamline,kim2026rethinking}. \Cref{fig:teaser} shows the question on
Qwen2.5-3B. We complement this work with a controlled comparison: for one pool of complete candidates and one recovery
recipe, which signal, measured when, orders the candidates by their downstream accuracy after full recovery, on
examples the signal has not seen?

We answer this on Qwen2.5-1.5B under full fine-tuning and LoRA and on Qwen2.5-3B under LoRA, and repeat the study on
Qwen2.5-0.5B (full fine-tuning) and on a second family, Llama-3.2-1B and 3B (LoRA): seven block-replacement searches
and two whole-layer-removal baselines at three latency budgets, every structure recovered with two seeds (three on
Llama-3.2-1B), and every task-based signal evaluated on examples disjoint from those of the final accuracy.
Our contributions are:
\begin{itemize}
\setlength{\itemsep}{1pt}
\item \textbf{A controlled comparison of selection signals across two model families.} The pre-recovery objective
separates strong structures from whole-layer removal but orders the strong ones significantly better than chance in
only one of six settings. After the same short recovery, zero-shot accuracy on held-out examples is the only signal
that is significant in all six settings; the divergence and the loss used by existing training-aware selection are
not.
\item \textbf{Different behaviour at the two Qwen sizes.} On the smaller model the results suggest a mismatch
between the objective and task performance: signals closer to the task decisions rank better, and zero-shot accuracy before recovery already orders the
structures, under full fine-tuning and LoRA alike. On the larger model none of the pre-specified pre-recovery signals
orders the structures, while a short recovery does. The Llama models do not show this failure, so we do not attribute
the difference to scale.
\item \textbf{A mechanism and its limit.} Stacking one block type makes the divergence grow faster than additive and
pairwise surrogates predict on both models, but the drop in task accuracy compounds only on the smaller model;
correcting the surrogate lowers the objective without improving the recovered model.
\item \textbf{Low observed regret, not higher accuracy.} No single search method is best everywhere; picking by
short-recovery zero-shot accuracy has the smallest mean and worst-case regret observed over the 18 setting--budget
cells, at a fraction of one recovery per candidate, and the zero-cost pre-recovery accuracy comes close on most models.
In prospective runs on three models that recover only the picks from pools of 16 near-optimal structures, it is
significantly worse than a searched structure in three of nine cells and better than all of them in one; the zero-step
rule, run on two of the models, is never significantly worse, which is not a proof of non-inferiority.
\end{itemize}

\section{Related Work}
\label{sec:related}
\textbf{Block replacement and layer removal.}
Puzzle~\citep{bercovich2024puzzle} and its vision predecessors~\citep{li2020dna,moons2021donna,molchanov2022lana}
distil alternative blocks and assemble a model by mixed-integer programming on additive per-block scores;
\citet{xu2026additivity} studied when this additivity breaks. Whole-layer removal ranks layers by input--output
similarity~\citep{men2025shortgpt}, by perplexity after removal~\citep{song2024sleb}, or by exact or pairwise
optimisation~\citep{jansen2026cbo,su2026repair}; \citet{gromov2025deeper} remove deep layers with little loss after
healing, and iterative removal accounts for interactions~\citep{chen2025prunecomp}. \citet{su2026repair} show that a
training-free repair changes which layers are best to remove, and \citet{kim2026rethinking} that the calibration
objective matters more than the search algorithm. We study heterogeneous block replacement and judge every signal
against the recovered model.

\textbf{Training-aware selection and selection signals.}
Minitron~\citep{muralidharan2024minitron} retrains a few width and depth configurations for about 400 steps and ranks
them by validation loss, observing that the ranking stabilises after about 300 steps. DarwinLM~\citep{tang2026darwinlm}
evolves per-layer sparsity, fine-tunes offspring on growing token budgets and selects by KL to the dense model; it shows
qualitatively that the KL after small-scale training predicts the KL after larger-scale training for 16 offspring of
Llama-2-7B. EvoPress~\citep{sieberling2025evopress} selects by KL on growing calibration samples without training.
Puzzle~\citep{bercovich2024puzzle} compares LM loss and KL as replace-one-block scores (KL gives better architectures)
and scores blocks on one half of MMLU to evaluate on the other, which helps the target task but costs general quality.
LOOP~\citep{xu2026loop} interleaves layer removal with short distillation, NIRVANA~\citep{ai2026nirvana} uses a
saliency meant to preserve fine-tuning dynamics, and TraceNAS~\citep{malettira2026tracenas} selects pruned models
without training by a gradient-based proxy of their potential after post-pruning training. For depth-pruned small
models, \citet{dhoka2026breaks} find that short pilot runs reveal which parameters are best to heal. Our study differs in its unit and target: complete candidate
structures from one fixed pool, ranked against their downstream accuracy after full recovery, with several signals at
several points of recovery, held-out task examples and permutation tests.

\textbf{Evaluation and proxies.}
Perplexity can stay unchanged while downstream ability drops~\citep{jaiswal2024llmkick}, and selecting layers by
perplexity can hurt downstream accuracy~\citep{chen2024streamline}. Architecture search uses training-free
proxies~\citep{abdelfattah2021zerocost}, successive halving~\citep{li2018hyperband} and learning-curve
predictors~\citep{white2021predictors}; here every candidate starts from pretrained weights and is recovered from the
same parent.

\section{Method}
\label{sec:method}
\textbf{Search space and objective.}
Each of the $L$ decoder layers keeps its block or takes one of seven replacements: half of the key--value head groups,
no attention, the MLP pruned to 50\% or 25\% of its hidden units, a low-rank MLP with half of the parameters, no MLP, or
no layer. Replacements with parameters are initialised from the parent layer and trained to reproduce its output
(3{,}000 steps). Latency is predicted from a per-option lookup table measured on the target GPU. The pre-recovery
objective is the token-level divergence $\KL(A)=\mathbb{E}_t\,\KL(p_{\text{parent}}\,\|\,p_A)$ over the full vocabulary
on 128 calibration sequences. It is optimised by the additive surrogate (ILP), a pairwise surrogate (IQP), a hybrid
search that measures structures directly~\citep{xu2026additivity}, the IQP with halved-attention layers capped at the
ILP's count (IQP-cap) or without halved-attention pair terms (IQP-noHH), and count-corrected surrogates
(Add-count, Pair-count; \cref{sec:count}). ShortGPT and SLEB remove whole layers under the same latency table
(\Sref{sec:setup}).

\textbf{Recovery.}
Every structure is recovered by distillation from the parent: 2{,}000 steps of 16 sequences of 512 tokens, two seeds.
Qwen2.5-1.5B is recovered by full fine-tuning and, separately, by LoRA~\citep{hu2022lora} (rank 64); Qwen2.5-3B by LoRA,
as full fine-tuning does not fit on one 32\,GB GPU. Quality is the macro zero-shot accuracy on ARC-easy and
ARC-challenge~\citep{clark2018arc}, HellaSwag~\citep{zellers2019hellaswag}, PIQA~\citep{bisk2020piqa} and
WinoGrande~\citep{sakaguchi2020winogrande}. Throughout this paper, full recovery denotes completion of the prescribed
2{,}000-step schedule, rather than convergence.

\textbf{Comparing signals.}
For every recovered structure a separate run repeats the first 400 steps of its recovery (same data order and
learning-rate schedule; under LoRA the adapter initialisation is drawn afresh) and records the calibration
KL, the distillation loss and the WikiText-2 perplexity~\citep{merity2017wikitext} at steps 0--400 and the zero-shot
accuracy at steps 100 and 400. A signal is judged by how it orders the structures of one budget relative to their
final accuracy (mean of the recovery seeds). We count \emph{decided pairs}, whose final accuracies differ by more than
0.5 points, and report the share ordered correctly. \emph{Common protocol} (\cref{tab:families}, used for all
comparisons between signals): the final accuracy is measured on one half of the examples of every task (odd or even
index), task-based signals on the other half, and signals that use no task data (KL, loss) unchanged; both split
directions are pooled, so every signal is scored on the same decided pairs. This is task-aware selection: the selection
half comes from the same five tasks; \Sref{sec:sens} also selects on four tasks and evaluates on the fifth.
Significance against chance is a within-budget permutation test that applies the same shuffle to both split
directions (20{,}000 shuffles, add-one $p$-values); a binomial test over pairs would be anti-conservative because pairs
share structures. Differences between two signals are tested by exchanging the two signals within each budget and
split direction (sign-flip randomisation over these units, \Sref{sec:sens}). The detailed Qwen tables in the appendix
and \cref{fig:signals} keep the protocol of the first version (KL-type signals against the final accuracy on all
examples), and the decomposition of \cref{sec:analysis} uses the examples it does not score; their pair counts differ.
Bootstrap intervals over examples are conditional on the recovery seeds; the seed standard deviation of a structure's
final accuracy is 0.17--0.28 points on average, against a spread of 1.1--2.7 points between the strong structures of a
budget.

\textbf{Two-stage selection.}
Following the select-after-short-training framework of Minitron and
DarwinLM~\citep{muralidharan2024minitron,tang2026darwinlm}, a pool of structures that meet the budget is recovered for
100 steps each and the one with the highest zero-shot accuracy on the
selection half is kept; only the signal differs from prior work.

\section{Experiments}
\label{sec:exp}
Parents are Qwen2.5-1.5B and 3B~\citep{yang2024qwen25}; calibration and training data come from disjoint
FineWeb-Edu documents~\citep{penedo2024fineweb}; budgets are 80, 70 and 60\% of the block latency on an RTX~5090.
Sections~\ref{sec:pre}--\ref{sec:signals} analyse these three settings in detail; \cref{sec:families} repeats the
signal comparison on Qwen2.5-0.5B and on Llama-3.2-1B and 3B. Details are in \Sref{sec:setup}.

\begin{table*}[t]
\centering
\caption{Zero-shot accuracy (\%, macro over five tasks) after 2{,}000 recovery steps, mean of two seeds. Bold: best
block-replacement structure of the budget. Parent: 1.5B 65.17, 3B 68.67.}
\label{tab:main}
\footnotesize\setlength{\tabcolsep}{4pt}
\begin{tabular}{llccccccccc}
\toprule
setting & budget & ILP & IQP & Hybrid & IQP-cap & IQP-noHH & Add-count & Pair-count & ShortGPT & SLEB \\
\midrule
1.5B full & 80\% & 59.27 & 59.47 & \textbf{60.40} & 60.33 & 60.24 & 58.40 & 59.04 & 57.81 & 58.02 \\
 & 70\% & 55.50 & 56.05 & 56.49 & \textbf{56.64} & 56.14 & 56.29 & 55.81 & 53.10 & 53.99 \\
 & 60\% & 51.15 & 52.25 & 51.69 & 52.32 & 52.03 & 52.38 & \textbf{52.41} & 47.91 & 50.47 \\
\midrule
1.5B LoRA & 80\% & 59.29 & 59.34 & \textbf{60.47} & 60.45 & 60.28 & 58.20 & 59.40 & -- & -- \\
 & 70\% & 54.76 & 55.66 & 56.37 & \textbf{56.47} & 55.86 & 55.55 & 55.02 & -- & -- \\
 & 60\% & 50.36 & 51.58 & 50.46 & \textbf{52.16} & 51.32 & 51.51 & 51.35 & -- & -- \\
\midrule
3B LoRA & 80\% & 65.10 & 63.41 & 65.34 & 64.31 & 65.28 & \textbf{65.97} & 64.75 & 60.28 & 61.83 \\
 & 70\% & 61.90 & 58.90 & \textbf{61.98} & 59.41 & 59.36 & 60.47 & 59.28 & 57.92 & 57.87 \\
 & 60\% & 56.16 & 55.33 & 56.09 & 54.82 & 54.42 & \textbf{56.88} & 54.78 & 50.26 & 52.86 \\
\bottomrule
\end{tabular}

\end{table*}

\subsection{Recovered accuracy}
Every block-replacement structure beats both whole-layer-removal baselines, by 0.4--4.5 points on the 1.5B model and
1.0--6.6 points on the 3B model (\cref{tab:main}). Among the block-replacement searches no method is best everywhere:
the best structure of a budget comes from four different methods across the nine setting--budget cells, and the seven searched
structures of a budget lie within 1.1--3.1 points of each other. Choosing among these strong structures is the
decision we study.

\subsection{The pre-recovery objective is not a reliable ranking of strong structures}
\label{sec:pre}
The calibration KL that the searches optimise orders 63\%, 61\% and 32\% of the decided pairs correctly on the three
settings under the common protocol (\cref{tab:families}; 67\%, 58\% and 31\% against all examples, \cref{tab:pre}),
none significantly better than chance; on the 3B model its observed ordering accuracy is below 50\%. It does
separate strong structures from whole-layer removal (84\% with the baselines on the 1.5B model). Zero-shot accuracy
before recovery behaves differently on the two models: on the 1.5B model it orders 88\% (full fine-tuning) and 82\%
(LoRA) of the decided pairs correctly, so a pre-recovery signal can rank strong structures when it measures the right
thing; on the 3B model it orders 32\%. The LoRA arm of the 1.5B model does not show the 3B behaviour, so the recovery
recipe alone does not explain it; with two model sizes we do not attribute it to scale.

\begin{table}[t]
\centering
\caption{Qwen, pre-recovery signals: decided pairs ordered correctly and permutation $p$ for agreement above chance.
Zero-shot accuracy uses held-out example halves; the KL is compared with the final accuracy on all examples (protocol of
the first version; \cref{tab:families} scores all signals on the same pairs).}
\label{tab:pre}
\scriptsize\setlength{\tabcolsep}{2pt}
\resizebox{\columnwidth}{!}{\begin{tabular}{lcc}
\toprule
setting & calibration KL & zero-shot accuracy \\
\midrule
1.5B full & 22/33 = 67\% ($p=0.130$) & 57/65 = 88\% ($p=0.0002$) \\
1.5B full (+ baselines) & 64/76 = 84\% ($p=0.0004$) & 137/152 = 90\% ($p=0.0002$) \\
1.5B LoRA & 26/45 = 58\% ($p=0.291$) & 72/88 = 82\% ($p=0.0002$) \\
3B LoRA & 15/48 = 31\% ($p=0.962$) & 29/92 = 32\% ($p=0.991$) \\
3B LoRA (+ baselines) & 58/92 = 63\% ($p=0.076$) & 95/180 = 53\% ($p=0.336$) \\
\bottomrule
\end{tabular}
}
\end{table}

\subsection{Which signal survives a short recovery}
\label{sec:signals}
\begin{figure*}[t]
\centering
\includegraphics[width=0.85\textwidth]{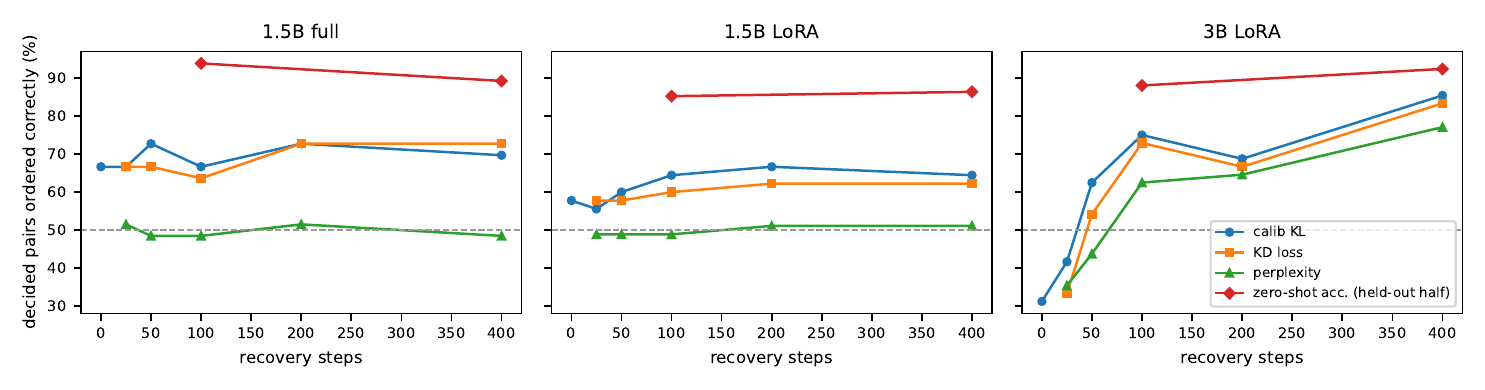}
\caption{Decided pairs ordered correctly by each signal along the first 400 recovery steps (step 0: before recovery;
dashed: chance).}
\label{fig:signals}
\end{figure*}

On the 3B model, where the pre-specified pre-recovery signals fail, 100 of the 2{,}000 recovery steps are enough for zero-shot
accuracy on held-out examples to order 88\% of the decided pairs correctly, against 76\% for the calibration KL and
74\% for the distillation loss at the same point; at 400 steps the shares are 92\%, 84\% and 83\% (common protocol,
\cref{tab:families}; \cref{tab:signals,fig:signals} give the detail of all checkpoints). On the 1.5B model the
divergence and the loss order 58--69\% of the pairs at 100 and 400 steps, significant only for the KL after 100 LoRA
steps ($p=0.044$), while zero-shot accuracy orders 85--94\% correctly.
Perplexity is the weakest signal. Zero-shot accuracy has the highest point estimate among the short-recovery signals at both
checkpoints on all three settings.

\begin{table*}[t]
\centering
\caption{Qwen, signals after 100 and 400 of the 2{,}000 recovery steps: decided pairs ordered correctly (permutation
$p$). Zero-shot accuracy on held-out example halves; KL, loss and perplexity against the final accuracy on all examples
(protocol of the first version; \cref{tab:families} scores all signals on the same pairs).}
\label{tab:signals}
\footnotesize\setlength{\tabcolsep}{6pt}
\begin{tabular}{llcccc}
\toprule
setting & steps & calib KL & KD loss & perplexity & zero-shot accuracy \\
\midrule
1.5B full & 100 & 67\% ($p=0.137$) & 64\% ($p=0.180$) & 48\% ($p=0.581$) & 94\% ($p=0.0002$) \\
 & 400 & 70\% ($p=0.088$) & 73\% ($p=0.066$) & 48\% ($p=0.581$) & 89\% ($p=0.0002$) \\
\midrule
1.5B LoRA & 100 & 64\% ($p=0.126$) & 60\% ($p=0.236$) & 49\% ($p=0.576$) & 85\% ($p=0.0002$) \\
 & 400 & 64\% ($p=0.125$) & 62\% ($p=0.173$) & 51\% ($p=0.510$) & 86\% ($p=0.0002$) \\
\midrule
3B LoRA & 100 & 75\% ($p=0.014$) & 73\% ($p=0.022$) & 62\% ($p=0.154$) & 88\% ($p=0.0002$) \\
 & 400 & 85\% ($p=0.001$) & 83\% ($p=0.002$) & 77\% ($p=0.008$) & 92\% ($p=0.0002$) \\
\bottomrule
\end{tabular}

\end{table*}

\subsection{A second model family and a third Qwen size}
\label{sec:families}
We repeated the whole pipeline on Qwen2.5-0.5B (full fine-tuning) and on Llama-3.2-1B and 3B (LoRA; three recovery seeds
on the 1B model), with the same searches, budgets, recovery recipe and statistics (\cref{tab:families}; structures
that two searches found identically are counted once). Under the common protocol, zero-shot accuracy after 100 recovery
steps is the only signal that orders the decided pairs significantly better than chance in all six settings
(80--94\%). The search objective is significant in none (32--72\%; 50\% on Llama-3.2-1B); the divergence and the
distillation loss after 100 or 400 steps, the signals of DarwinLM and Minitron, are significant in two or three of the
six settings. Pre-recovery zero-shot accuracy is significant everywhere except Qwen2.5-3B, so the failure of the cheap
task signal on that model is not shared by the Llama models of similar size. At 400 steps zero-shot accuracy is not
significant on the 0.5B model (67\%), so more short-recovery steps do not always help.

Significance against chance does not by itself show that one signal is better than another, so we also compare the
signals directly (\Sref{sec:sens}). Pooled over the six settings, zero-shot accuracy after 100 steps orders 13--19
points more of the same decided pairs correctly than every divergence or loss signal after 100 or 400 steps, 30 points
more than the search objective and 14 more than pre-recovery zero-shot accuracy (sign-flip $p\le0.01$); it is not
distinguishable from zero-shot accuracy after 400 steps (+1 point). Within a single setting the test has only six units
and is significant only in a few comparisons, and the pooled units share structures across the two split directions,
so the pooled $p$-values are approximate. The ranking result holds when the decided-pair threshold is 1 point (all six
settings) and when the candidates are selected on four tasks and evaluated on the fifth (all six settings, 68--79\%);
counting every pair (threshold 0) it holds in five of six. The search objective is significant in one setting at
each of the other thresholds (Qwen2.5-0.5B counting every pair, 69\%; Llama-3.2-3B at 1 point, 80\%), so its failure is
not uniform but its ranking is inconsistent across settings and thresholds.

\begin{table*}[t]
\centering
\caption{Six settings, common protocol: decided pairs ordered correctly (\%) and within-budget permutation $p$
(20{,}000 joint shuffles, add-one) among the seven searched structures of every budget (identical structures merged).
Final accuracy on one half of every task's examples, task-based signals on the other half, both directions pooled, so
every signal is scored on the same pairs. Bold: $p<0.05$. Llama-3.2-1B: mean of three recovery seeds.}
\label{tab:families}
\footnotesize\setlength{\tabcolsep}{4pt}
\begin{tabular}{llcccccc}
\toprule
 & & Qwen 0.5B & Qwen 1.5B & Qwen 1.5B & Qwen 3B & Llama 1B & Llama 3B \\
Signal & When & full & full & LoRA & LoRA & LoRA & LoRA \\
\midrule
Calibration KL (search objective) & before & 72 (.076) & 63 (.17) & 61 (.17) & 32 (.96) & 50 (.53) & 68 (.064) \\
Zero-shot accuracy & before & \textbf{80 (.011)} & \textbf{88 ($<$.001)} & \textbf{82 (.001)} & 32 (.96) & \textbf{77 (.022)} & \textbf{85 ($<$.001)} \\
\addlinespace
Calibration KL & 100 steps & 74 (.052) & 62 (.20) & \textbf{69 (.044)} & \textbf{76 (.007)} & 66 (.17) & \textbf{72 (.032)} \\
Distillation loss & 100 steps & 74 (.052) & 58 (.27) & 65 (.10) & \textbf{74 (.014)} & 59 (.31) & \textbf{70 (.040)} \\
Zero-shot accuracy & 100 steps & \textbf{82 (.008)} & \textbf{94 ($<$.001)} & \textbf{85 ($<$.001)} & \textbf{88 ($<$.001)} & \textbf{84 (.009)} & \textbf{80 (.002)} \\
\addlinespace
Calibration KL & 400 steps & \textbf{77 (.034)} & 65 (.14) & 68 (.056) & \textbf{84 ($<$.001)} & 66 (.17) & \textbf{75 (.013)} \\
Distillation loss (Minitron-style) & 400 steps & \textbf{77 (.034)} & 68 (.092) & 66 (.082) & \textbf{83 (.001)} & 66 (.17) & \textbf{74 (.018)} \\
Zero-shot accuracy & 400 steps & 67 (.10) & \textbf{89 ($<$.001)} & \textbf{86 ($<$.001)} & \textbf{92 ($<$.001)} & \textbf{80 (.021)} & \textbf{84 ($<$.001)} \\
\midrule
Decided pairs & & 39 & 65 & 88 & 92 & 44 & 81 \\
\bottomrule
\end{tabular}

\end{table*}

\subsection{Selecting among the searched structures}
\label{sec:select}
Applied as a selection rule to the seven searched structures of every budget (\cref{tab:twostage}), zero-shot accuracy
after 100 steps gives the highest mean final accuracy of the rules on all three settings; its mean regret to the best
structure of the pool is 0.05 points on the 3B model and 0.16 and 0.58 points on the 1.5B model with full fine-tuning
and LoRA, and it is above always deploying the hybrid search's structure. Picking by the pre-recovery KL is worse than
deploying the additive ILP on the 3B model. One 100-step run with a zero-shot evaluation costs 10--12\% of a full
recovery, so the rule costs 0.7--0.8 full recoveries for a pool of seven candidates, in addition to recovering the
chosen one.

\textbf{The best search changes from cell to cell.} Over all six settings and three budgets (18 cells), the best of
the seven searched structures comes from six different methods, so committing to one search method in advance has a
cost (\cref{tab:regret}). Always deploying the hybrid search, the best fixed choice, falls 0.47 points behind the best
structure on average and up to 1.70 points; always deploying the ILP falls up to 3.38 points behind, and picking by the
search objective up to 2.85. Picking by zero-shot accuracy after 100 steps has the smallest mean (0.27) and worst-case
(0.81) regret and is more than 0.5 points behind in only three cells. Pre-recovery zero-shot accuracy, which needs no
recovery at all, is second (0.39, worst 1.21, on Qwen2.5-3B). The short recovery therefore buys 0.12 points of mean
and 0.40 points of worst-case regret over the zero-step rule, at 0.7--0.8 full recoveries for seven candidates against
seven zero-shot evaluations; whether that is worth it depends on how costly a wrong deployment is. In the prospective runs the zero-step rule
did at least as well as the 100-step rule. The regrets are
averaged over the two example halves of a cell and are descriptive (cells share structures).

\begin{table}[t]
\centering
\caption{Regret (points of final accuracy behind the best of the seven searched structures of a cell) over 18 cells
(six settings $\times$ three budgets): mean, worst cell, and number of cells with regret above 0.5 and 1 point.
Task-based rules select on one half of the examples and are evaluated on the other (both directions averaged).
Descriptive: cells share structures and are not independent.}
\label{tab:regret}
\scriptsize\setlength{\tabcolsep}{3pt}
\begin{tabular}{lcccc}
\toprule
Rule & Mean & Worst & $>$0.5 & $>$1 \\
\midrule
Always the additive ILP & 1.27 & 3.38 & 15 & 9 \\
Always the pairwise IQP & 0.95 & 3.23 & 12 & 7 \\
Always the hybrid search & 0.47 & 1.70 & 8 & 1 \\
Always Add-count & 1.03 & 2.54 & 12 & 10 \\
Always Pair-count & 1.11 & 2.85 & 15 & 10 \\
\midrule
Pre-recovery calibration KL & 1.11 & 2.85 & 14 & 9 \\
Pre-recovery zero-shot accuracy & 0.39 & 1.21 & 8 & 1 \\
\midrule
Calibration KL, 100 steps & 0.63 & 1.41 & 11 & 5 \\
Distillation loss, 100 steps & 0.63 & 1.41 & 11 & 5 \\
Distillation loss, 400 steps & 0.57 & 1.47 & 10 & 4 \\
\textbf{Zero-shot accuracy, 100 steps} & \textbf{0.27} & \textbf{0.81} & \textbf{3} & \textbf{0} \\
Zero-shot accuracy, 400 steps & 0.31 & 1.02 & 5 & 1 \\
\bottomrule
\end{tabular}

\end{table}

\textbf{Prospective test.} The comparison above is retrospective: the rules pick among structures that were all
recovered. To test the rule prospectively on the 3B model, we build a pool of 16 distinct near-optimal structures per
budget by solving the count-corrected additive surrogate repeatedly with no-good cuts (members differ in at least
two layers; member 0 is the surrogate optimum that a one-stage search would deploy), run the 100-step
proxy recovery for every member, let each rule pick one member, and fully recover only the picks (two seeds). Final
accuracy is measured on the odd-indexed examples; the task-based rule selects on the even-indexed ones
(\cref{tab:prospective}). Zero-shot accuracy after 100 steps gives the highest mean final accuracy of the rules
(61.36), but the surrogate optimum of the pool, which needs no proxy recovery, gives the same mean (61.35): both pick
the same structure at 60\%, and they differ by 0.24 and $-0.22$ points at 80\% and 70\%. Against the searched
structures the picks depend on the budget. At 60\% the rule is better than the ILP, the IQP and the hybrid (by
1.85, 1.95 and 1.39 points; paired bootstrap 95\% intervals exclude zero). At 70\% every rule's pick is worse than the
hybrid's structure (by 1.43--1.65 points, intervals exclude zero), and the zero-shot pick is also worse than the
ILP's. At 80\% it is level with the ILP and the hybrid. The divergence after 100 steps and the pre-recovery KL are the
weakest rules on this pool, as in the retrospective comparison. In this run the short-recovery zero-shot rule had low regret among the candidates but did not add accuracy over a
well-built surrogate pool, and no rule recovered the hybrid's advantage at 70\%.

We repeated the prospective run on Llama-3.2-1B with three recovery seeds. The zero-shot pick is within the paired
bootstrap intervals of the ILP, the IQP and the hybrid at all three budgets (differences $-0.87$ to $+0.35$ points).
Picking by the search objective, which deploys the pool's surrogate optimum at 80 and 70\%, is significantly worse
than the ILP and the IQP at 70\% (by 1.10 and 1.80 points), and the divergence and loss after 100 steps and the
concentration rule are significantly worse than the IQP there; at 60\% the KL- and loss-based picks are significantly
better than the ILP (by 1.14 points). With the ILP, IQP and hybrid structures added to the pool, the zero-shot rule picks
the hybrid, the IQP and a pool member at the three budgets and is again never significantly worse than any searched
structure. On Llama-3.2-1B it is the only one of these six rules that was never significantly below a searched structure; on
Qwen2.5-3B every rule fell below the hybrid at 70\%, and with the searched structures added to that pool the zero-shot
rule was never significantly below the hybrid (\Sref{sec:aug}). Only on Qwen2.5-3B at 60\% was a pick significantly better than all three
searched structures, and that pick is the pool's surrogate optimum.

\textbf{A third model and the zero-step rule.} We repeated the prospective run on Qwen2.5-0.5B with full fine-tuning
(two seeds) and, on Llama-3.2-1B and Qwen2.5-0.5B, added the zero-step rule: pick the pool member with the highest
pre-recovery zero-shot accuracy on the even-indexed examples (\cref{tab:prosp2}). On Qwen2.5-0.5B the 100-step
zero-shot rule does poorly: it is significantly below the IQP at 80\% (by 1.08 points) and below the ILP and the hybrid
at 70\% (by 1.28 and 1.86 points; it picks the surrogate optimum there), and significantly above the ILP and the hybrid
at 60\% (by 1.96 and 1.53). The zero-step rule is never significantly below a searched structure on either model. On
Qwen2.5-0.5B it is 0.98 points above the 100-step rule at 70\% (interval excludes zero) and significantly above the ILP
and the hybrid at 60\% (by 1.87 and 1.44), and its mean over the budgets is above that of the 100-step rule on both
models (50.83 against 50.57 on Llama-3.2-1B, 49.00 against 48.58 on Qwen2.5-0.5B). Over the three models the 100-step
rule is significantly below a searched structure in three of nine cells, so its retrospective advantage over the
zero-step rule (\cref{tab:regret}) did not carry over to the prospective pools. The zero-step rule was not run on
Qwen2.5-3B, where pre-recovery zero-shot accuracy does not order the searched structures (\cref{tab:families}).

\begin{table}[t]
\centering
\caption{Prospective two-stage selection on Qwen2.5-3B (LoRA): final accuracy (\%, odd-indexed half, mean of two
seeds) of the searched structures and of the pool member picked by each rule. Bold: best of the budget.
$^\dagger$Exploratory statistic of \cref{sec:analysis}, fixed in advance as a rule.}
\label{tab:prospective}
\scriptsize\setlength{\tabcolsep}{2pt}
\resizebox{\columnwidth}{!}{\begin{tabular}{lcccc}
\toprule
structure & 80\% & 70\% & 60\% & mean \\
\midrule
searched: ILP & 64.90 & 62.20 & 55.75 & 60.95 \\
searched: IQP & 62.85 & 58.98 & 55.65 & 59.16 \\
searched: Hybrid & 65.18 & \textbf{62.66} & 56.21 & 61.35 \\
\midrule
pool: surrogate optimum (no recovery) & 65.23 & 61.23 & \textbf{57.59} & 61.35 \\
pool: pre-recovery calib.\ KL (min) & 64.18 & 61.23 & 56.54 & 60.65 \\
pool: concentration$^\dagger$ (max) & \textbf{65.47} & 61.23 & 56.84 & 61.18 \\
pool: KL after 100 steps (min) & 63.91 & 61.23 & 56.84 & 60.66 \\
pool: KD loss after 100 steps (min) & 65.23 & 61.23 & 56.36 & 60.94 \\
pool: zero-shot acc.\ after 100 steps (even half) & \textbf{65.47} & 61.01 & \textbf{57.59} & \textbf{61.36} \\
\bottomrule
\end{tabular}
}
\end{table}

\begin{table}[t]
\centering
\caption{Final accuracy (\%, held-out half) of the structure picked by each rule among the seven searched structures,
mean over budgets and split directions.}
\label{tab:twostage}
\scriptsize\setlength{\tabcolsep}{2pt}
\resizebox{\columnwidth}{!}{\begin{tabular}{lccc}
\toprule
selection rule & 1.5B full & 1.5B LoRA & 3B LoRA \\
\midrule
fixed: ILP & 55.31 & 54.81 & 61.05 \\
fixed: Hybrid & 56.19 & 55.77 & 61.14 \\
pre-recovery calib.\ KL (min) & 55.75 & 55.26 & 59.33 \\
KL after 100 steps (min) & 55.75 & 55.74 & 61.40 \\
KD loss after 100 steps (min) & 55.75 & 55.74 & 61.40 \\
zero-shot acc.\ after 100 steps (held-out half) & 56.40 & 55.96 & 61.61 \\
oracle & 56.56 & 56.54 & 61.66 \\
\bottomrule
\end{tabular}
}
\end{table}

\section{Analysis}
\label{sec:analysis}
\textbf{Why the divergence misses what matters.}
We decompose the pre-recovery divergence of every structure, token by token, and test each part as a ranking signal
against the final accuracy on task examples not used by the decomposition (\cref{tab:decomp}). On the 1.5B model
task-oriented signals generally rank better than the mean calibration KL: the calibration KL (64\%,
59\%) to the KL on task answer tokens (79\%, 78\%), the agreement of option decisions with the parent (82\%, 76\%) and
the zero-shot accuracy itself (88\%, 83\%). The KL on the tokens the parent is most confident about shows no significant ranking
ability (55\%, 54\%). This suggests that the search objective averages over many tokens whose divergence matters
little for the decisions that the evaluation scores. On the 3B model every one of these quality measures orders fewer than 40\% of
the decided pairs correctly before recovery.

\textbf{Exploratory: concentration of the divergence.}
One statistic of the decomposition, which was not part of our planned comparison, ranks the structures on all three
settings, including the 3B model: the share of the total divergence carried by the 10\% of tokens with the largest
divergence. Structures whose divergence is concentrated on few tokens end up more accurate (88\%, 85\% and 81\% of the
decided pairs; $p\le0.003$, and $p<0.05$ after Holm adjustment over the 14 statistics we tested). The share is nearly
uncorrelated with the mean divergence (rank correlation $\le0.07$) and with the number of replaced blocks, keeps a
partial rank correlation of 0.56--0.72 with the final accuracy when both are controlled for, holds when any single
search method is left out (73--96\%) and orders the whole-layer-removal baselines as well (90--92\% with them). A
plausible reading is that recovery repairs damage confined to few tokens more easily than damage spread over many.
Token-level damage also guides ReCal~\citep{jiang2026recal}, which uses the teacher--student KL to find heavily
affected tokens and reweights the pruning calibration to preserve them, reporting gains that persist through
recovery; our statistic instead describes how the divergence of a given structure is distributed over tokens, is
nearly independent of its mean, and is used to select among complete structures.
It also holds for every other concentration statistic we computed afterwards (top-5\%, top-20\%, Gini, entropy;
\Sref{sec:conc}). Because it was found after the fact, we treat it as a hypothesis and test it, with the criterion
fixed in advance, as a selection rule and on new models. As a rule in the prospective run (\cref{tab:prospective}) it
picks the same structure as the zero-shot rule at 80\%, is level with the other rules at 70\% and is 0.75 points below
the zero-shot pick at 60\% (mean 61.18 against 61.36), so it is competitive without any recovery but not better.
Out of sample, with the statistic and its direction fixed, it replicates on Llama-3.2-1B (83\% of 24 decided pairs,
$p=0.009$; 81\%, $p=0.03$ with the two seeds available at the time of the test) but not on Llama-3.2-3B (59\% of 41,
$p=0.25$), where the mean divergence itself does slightly better (68\%, $p=0.07$). One success in two independent tests
leaves it an open hypothesis rather than a selection signal.

\textbf{A pre-specified diagnostic.}
Pre-recovery zero-shot accuracy ranks the structures on the 1.5B model but not on the 3B model, so a practitioner needs
to know in advance whether the cheap signal can be trusted. On the nine Qwen cells (model $\times$ recipe $\times$
budget) the rank correlation between the calibration KL of the seven searched structures before recovery and after 25
recovery steps separates the two cases: 0.89--1.00 on the 1.5B model and 0.79--0.86 on the 3B model. We fixed a rule
from these cells (threshold 0.875: above it, select by pre-recovery zero-shot accuracy; below it, by zero-shot
accuracy after 100 steps), which gives a mean regret of 0.14 points at 0.58 of the cost of running the short recovery
for every candidate, and fixed the threshold and predictions in a time-stamped file with recorded hashes before the
proxy runs of new models (\Sref{sec:prereg}). The file was not deposited publicly at the time, so the hashes verify its
content but not its date; we therefore call it pre-specified rather than pre-registered. The first test, on Llama-3.2-1B, predicted that all three budgets are stable. The
statistic is 0.991, 0.857 and 0.964 at 80, 70 and 60\%, so two of three cells meet the threshold, a partial pass by the
pre-specified criterion. At 70\% three structures have almost equal pre-recovery KL (differences below 0.001), and their
reordering after 25 steps lowers the statistic. This reading was made after seeing the data, so we report the cell as a
failure. Because pre-recovery zero-shot accuracy was already known to work on Llama-3.2-1B, this test checks only the
diagnostic side. The test of prediction 3b on Llama-3.2-3B, the first model run after the specification, is one-sided by
outcome: the statistic is 1.000, 0.883 and 1.000, so all three cells are classified stable (the 70\% cell by a margin
of 0.008), and in these cells pre-recovery zero-shot accuracy orders 35 of 43 decided pairs correctly (81\%; even-index
selection, odd-index evaluation), meeting the specified $\ge75\%$. With no unstable cell, the specified prediction
for unstable cells ($\le60\%$) could not be tested, so the diagnostic remains unconfirmed as a method; following it
costs 0.41 points of regret on this model at a quarter of the proxy cost.

\begin{table*}[t]
\centering
\caption{Pre-recovery decomposition of the divergence as ranking signals: decided pairs ordered correctly (permutation
$p$), final accuracy on task examples not used by the decomposition. $^\dagger$Exploratory: found after the fact; higher
share ranks higher.}
\label{tab:decomp}
\footnotesize\setlength{\tabcolsep}{6pt}
\begin{tabular}{lccc}
\toprule
pre-recovery signal & 1.5B full & 1.5B LoRA & 3B LoRA \\
\midrule
calibration KL (search objective) & 64\% (0.192) & 59\% (0.272) & 30\% (0.973) \\
\quad on the 25\% most confident parent tokens & 55\% (0.417) & 54\% (0.419) & 34\% (0.938) \\
\quad on the 25\% least confident parent tokens & 73\% (0.053) & 66\% (0.125) & 23\% (0.995) \\
KL on task answer tokens & 79\% (0.019) & 78\% (0.014) & 21\% (0.997) \\
option decisions shared with the parent & 82\% (0.010) & 76\% (0.024) & 21\% (0.998) \\
zero-shot accuracy & 88\% (0.002) & 83\% (0.002) & 26\% (0.992) \\
concentration: share of KL in the top 10\% tokens$^\dagger$ & 88\% (0.002) & 85\% (0.003) & 81\% (0.003) \\
\bottomrule
\end{tabular}

\end{table*}

\textbf{Stacking compounds the divergence.}
Replacing one block type in 16 layers, the additive surrogate underestimates the measured KL for all seven types on
the 1.5B model and six of seven on the 3B model, by up to 63\%. Measured pair interactions are negative for halved
attention, so their sum, which grows with the square of the number of replaced layers, predicts a negative divergence
for 16 halved layers on both models, while triples have positive residuals of 6--9\%; on the 3B model the pairwise
prediction underestimates four of seven types by about 50\% or more (\Sref{sec:stackdetail}).

\textbf{Compounding in the divergence need not reach the task.}
\citet{gromov2025deeper} observe that removing deep layers raises perplexity without a matching loss of question
answering accuracy. We repeat the stacking probe for three block types and measure, without recovery, the zero-shot
accuracy on the first 500 examples of every task, comparing the drop of the stacked model with the sum of the
single-layer drops (\cref{tab:stacktask}). On the 1.5B model the accuracy drop compounds like the divergence: for 16
layers the measured drop exceeds the additive one for all three types (12.4 against 7.2 points for skipped attention,
6.9 against 1.2 for the pruned MLP). On the 3B model the divergence still compounds (0.29 measured against 0.13 additive
for skipped attention), but the accuracy drop does not: it is close to additive for skipped attention (4.0 against
3.2) and well below additive for halved attention (0.5 against 5.6) and the pruned MLP (4.2 against 6.6). On the 3B
model the divergence therefore overstates the task damage of stacked replacements, consistent with its failure as a
ranking signal on that model.

\textbf{A better surrogate is not enough.}
\label{sec:count}
A per-type stacking term $\sum_t b_t\,n_t(n_t-1)/2$, fitted on measured prefixes of searched structures and entered
exactly into the integer program, halves the surrogate's error on large structures (\Sref{sec:surrogate}). On the 3B
model the corrected pairwise search finds structures with a lower pre-recovery KL than the hybrid at all three
budgets, yet after recovery it is significantly worse than the ILP and the hybrid at 70\% and 60\%, while the corrected
additive search, with a higher pre-recovery KL, is best at two budgets (\Sref{sec:countdetail}). Optimising the
pre-recovery objective more accurately does not select better recovered models.

\textbf{Robustness.} Using the same task examples for selection and evaluation would raise the share of correctly
ordered pairs by 5--10 points (\Sref{sec:split}); all numbers above use disjoint halves. One proxy run diverged; two
reruns leave every statistic unchanged (\Sref{sec:diverge}). Repeating all 27 proxy runs of Llama-3.2-1B with a
different seed (data order and adapter initialisation) leaves the zero-shot accuracy after 100 steps almost unchanged:
rank correlation 0.99 with the first runs over the 27 structures (0.83--1.00 within budgets), mean absolute difference
0.15 points. Recovering the same 0.5B structure four times (two searches found it) spread its final accuracy over 0.8
points, about the size of the differences between strong structures.

\section{Conclusion}
\label{sec:conclusion}
For block-replacement compression of Qwen2.5-0.5B, 1.5B and 3B and Llama-3.2-1B and 3B, the objective that searches
optimise before recovery did not order strong candidates by their recovered accuracy significantly better than chance
in any of six settings. On Qwen2.5-1.5B
the results suggest a mismatch between the mean calibration KL and downstream task performance; on Qwen2.5-3B none of the
pre-specified pre-recovery signals ordered the candidates, while zero-shot evaluation on held-out examples after a
short recovery did. After the same short recovery, zero-shot accuracy was the only signal significantly better than chance in all six
settings, unlike the divergence and the loss used by existing training-aware selection, and as a selection rule it
had the smallest worst-case regret because the best search method changes from cell to cell. Tested prospectively on
three models, it was significantly worse than a searched structure in three of nine cells and better than all of them
in one, while the zero-step rule, run on two of the models, was never significantly worse; on Qwen2.5-3B it did not improve on
the surrogate optimum of a pool of near-optimal structures. On Qwen2.5-3B the divergence compounds when one block type is stacked while the task
accuracy does not. An exploratory statistic, the
concentration of the divergence on few tokens, ranked the candidates on both Qwen models without any recovery,
replicated on Llama-3.2-1B and failed on Llama-3.2-3B.

\textbf{Limitations.} Two model families with models up to 3B parameters, so differences between sizes are
observations, not a scale trend; a 7B model is in progress. The 3B models are recovered with LoRA only, although the
LoRA arm of the 1.5B model does not reproduce the Qwen2.5-3B behaviour. The detailed analyses (decomposition,
stacking, surrogates) cover the Qwen models; the Llama models enter the signal comparison, the regret analysis, the
prospective test and the pre-specified tests. The threshold of the
pre-specified diagnostic was chosen on nine cells of two models, and the gap between its stable and unstable cells is
about 0.03. The prospective runs cover three models and one pool construction, and the zero-step rule was run on two of them. The ranking statistics rest on 39--92 decided pairs per
setting under the common protocol; the detailed Qwen tables and the decomposition use other example sets, as stated
in their captions. Paired comparisons between signals have few units per setting, and the pooled test treats units
that share structures as exchangeable. Selection is task-aware: it uses examples of the evaluation tasks (one half, or
four of the five tasks); selection on the tasks' training splits or on unrelated tasks is not tested. Two or three
recovery seeds per structure and intervals conditional on them do not capture all retraining variance. The stacking test of task accuracy is measured
before recovery on three block types. The exploratory concentration statistic still awaits confirmation on new models.

\section*{Impact Statement}
This paper presents work whose goal is to advance the field of Machine Learning. There are many potential societal
consequences of our work, none which we feel must be specifically highlighted here.

\section*{Use of Generative AI}
The authors used an AI assistant (Claude, Anthropic) to help design the experiments, write and review the
experiment code, operate the experiment pipelines, carry out the statistical analyses and draft the text. Every
number in the paper is produced by scripts from the raw outputs of the experiments; the authors inspected these
outputs and checked the text against the generated tables. The authors
take full responsibility for the content.

\bibliography{refs}
\bibliographystyle{icml2026}

\newpage
\appendix
\onecolumn
\begin{center}{\Large\bfseries Supplementary Material}\end{center}

\section{Setup details}
\label{sec:setup}
\paragraph{Parents and data.} Qwen2.5-1.5B (28 layers) and Qwen2.5-3B (36 layers), bfloat16; for
\cref{sec:families} also Qwen2.5-0.5B (24 layers) and Llama-3.2-1B (16 layers) and 3B (28 layers). Llama data are
tokenised with the Llama tokenizer and a beginning-of-sequence token; everything else is identical. Parent zero-shot
accuracy: Qwen2.5-0.5B 54.15, Llama-3.2-1B 59.99, Llama-3.2-3B 68.31. Calibration: 128
sequences of 512 tokens from the first 2{,}000 documents of a FineWeb-Edu shard; training and recovery data: 80{,}000
sequences from the remaining documents (41M tokens). WikiText-2 test in 512-token chunks. Zero-shot tasks: ARC-easy,
ARC-challenge, HellaSwag, PIQA (length-normalised option log-likelihood) and WinoGrande (partial scoring), full
evaluation sets.

\paragraph{Library.} Halved attention keeps the key--value head groups with the largest product of value and output
projection norms, together with their query heads; MLP pruning keeps the hidden units with the largest product of
gate, up and down projection norms; the low-rank MLP uses truncated SVD of each projection with half of the
parameters. Every replacement with parameters is trained for 3{,}000 steps (batch 8 sequences, AdamW, learning rate
$3\times10^{-4}$, 100 warm-up steps, cosine) on the normalised mean squared error between its output and the parent
layer's output, given the parent layer's input.

\paragraph{Latency.} Per-option block latency measured on an RTX~5090 for 8 sequences of 512 tokens (prefill, median
of 100 runs after 20 warm-up runs); the sum of block latencies predicts the latency of 20 random structures with a
mean absolute error of 1.6\% (1.5B) and 0.3\% (3B).

\paragraph{Searches.} Replace-one-block scores for every layer and option; pairwise scores for every option pair of
layers at most two apart and for 8 random option pairs of every farther layer pair (5{,}197 pairs for the 1.5B and 7{,}869 for the 3B model); integer
programs solved with HiGHS~\cite{huangfu2018highs}. The hybrid search ranks one- and two-layer moves with a summed-logit
predictor and measures the best on a quarter of the calibration set~\cite{xu2026additivity}. ShortGPT ranks layers by
$1-\cos$ of their input and output hidden states; SLEB greedily removes the layer whose removal gives the lowest
calibration perplexity on a quarter of the calibration set.

\paragraph{Recovery.} Global distillation (token-level KL to the parent over the full vocabulary), 2{,}000 steps of 16
sequences of 512 tokens, AdamW (weight decay 0.05), 100 warm-up steps and cosine decay, seeds 0 and 1. Full
fine-tuning of the 1.5B model: learning rate $5\times10^{-5}$, embedding and tied output layer frozen, activation
checkpointing. LoRA: rank 64, $\alpha=64$, on every linear layer of the decoder layers (including replaced ones),
learning rate $2\times10^{-4}$.

\paragraph{Further models.} Qwen2.5-0.5B is recovered by full fine-tuning (learning rate $5\times10^{-5}$,
embeddings trained) and Llama-3.2-1B and 3B by LoRA as above, all with the same 2{,}000-step schedule. Llama-3.2-1B was
run on an RTX~5070~Ti with its own latency table; Qwen2.5-0.5B and Llama-3.2-3B on the RTX~5090. Llama-3.2-1B has a
third recovery seed for every structure of \cref{tab:families} and of its prospective run. The prospective run of
Qwen2.5-0.5B (pool and budgets from the RTX~5090 latency table) was recovered on the RTX~5070~Ti.

\paragraph{Compute.} One recovery takes about 40 (1.5B) and 53 (3B) minutes on the RTX~5090; one 100-step proxy run
with a zero-shot evaluation about 4 and 6 minutes.

\begin{table}[H]
\centering
\caption{Pre-recovery calibration KL of every structure (the search objective).}
\label{tab:prekl}
\footnotesize\setlength{\tabcolsep}{3pt}
\begin{tabular}{llccccccccc}
\toprule
model & budget & ILP & IQP & Hybrid & IQP-cap & IQP-noHH & Add-count & Pair-count & ShortGPT & SLEB \\
\midrule
1.5B & 80\% & 0.339 & 0.306 & 0.290 & 0.294 & 0.321 & 0.292 & 0.276 & 1.015 & 0.390 \\
 & 70\% & 0.732 & 0.523 & 0.534 & 0.554 & 0.588 & 0.561 & 0.509 & 1.747 & 0.727 \\
 & 60\% & 1.070 & 0.827 & 0.989 & 0.868 & 0.861 & 0.842 & 0.785 & 3.282 & 1.299 \\
\midrule
3B & 80\% & 0.284 & 0.359 & 0.270 & 0.239 & 0.286 & 0.325 & 0.254 & 0.764 & 0.384 \\
 & 70\% & 0.682 & 0.552 & 0.484 & 0.521 & 0.609 & 0.568 & 0.480 & 1.017 & 0.680 \\
 & 60\% & 1.231 & 0.964 & 0.864 & 0.826 & 0.754 & 0.889 & 0.795 & 2.174 & 1.289 \\
\bottomrule
\end{tabular}

\end{table}

\section{Per-task results}
\label{sec:pertask}
Per-task zero-shot accuracy (\%, mean of two recovery seeds) and WikiText-2 perplexity after recovery
(Tables~\ref{tab:pt15f}--\ref{tab:pt3b}).

\begin{table}[H]\centering\caption{Qwen2.5-1.5B, full fine-tuning.}\label{tab:pt15f}\footnotesize
\begin{tabular}{llccccccc}
\toprule
budget & structure & ARC-e & ARC-c & HellaSwag & PIQA & WinoG. & macro & ppl \\
\midrule
80\% & ILP & 68.5 & 38.4 & 58.4 & 73.5 & 57.5 & 59.27 & 15.62 \\
80\% & IQP & 67.3 & 39.2 & 58.8 & 73.4 & 58.6 & 59.47 & 15.74 \\
80\% & Hybrid & 69.7 & 39.5 & 59.8 & 73.4 & 59.6 & 60.40 & 15.67 \\
80\% & IQP-cap & 70.3 & 39.8 & 59.2 & 73.4 & 58.9 & 60.33 & 15.22 \\
80\% & IQP-noHH & 68.5 & 39.2 & 59.6 & 74.0 & 59.9 & 60.24 & 15.83 \\
80\% & Add-count & 65.0 & 36.8 & 58.3 & 73.8 & 58.1 & 58.40 & 15.21 \\
80\% & Pair-count & 67.6 & 37.9 & 58.9 & 73.8 & 57.0 & 59.04 & 14.82 \\
80\% & ShortGPT & 65.8 & 36.9 & 56.7 & 71.4 & 58.2 & 57.81 & 17.54 \\
80\% & SLEB & 64.8 & 37.3 & 57.9 & 74.3 & 55.8 & 58.02 & 14.48 \\
\midrule
70\% & ILP & 63.6 & 34.0 & 51.8 & 70.2 & 57.9 & 55.50 & 18.86 \\
70\% & IQP & 63.7 & 35.2 & 53.3 & 71.5 & 56.6 & 56.05 & 17.48 \\
70\% & Hybrid & 63.7 & 34.8 & 53.8 & 71.8 & 58.4 & 56.49 & 18.17 \\
70\% & IQP-cap & 64.9 & 36.7 & 54.1 & 72.6 & 54.9 & 56.64 & 17.71 \\
70\% & IQP-noHH & 64.1 & 34.6 & 53.5 & 72.3 & 56.3 & 56.14 & 18.34 \\
70\% & Add-count & 64.9 & 36.5 & 53.1 & 71.9 & 55.1 & 56.29 & 17.98 \\
70\% & Pair-count & 63.6 & 34.8 & 53.2 & 71.6 & 55.8 & 55.81 & 17.11 \\
70\% & ShortGPT & 58.6 & 32.1 & 51.0 & 68.8 & 55.1 & 53.10 & 20.30 \\
70\% & SLEB & 60.7 & 32.9 & 50.8 & 72.2 & 53.3 & 53.99 & 16.46 \\
\midrule
60\% & ILP & 56.9 & 30.5 & 45.0 & 68.4 & 55.0 & 51.15 & 23.25 \\
60\% & IQP & 58.3 & 31.4 & 47.0 & 70.6 & 53.9 & 52.25 & 20.19 \\
60\% & Hybrid & 58.3 & 30.8 & 46.1 & 68.9 & 54.4 & 51.69 & 23.07 \\
60\% & IQP-cap & 59.8 & 30.6 & 47.0 & 69.4 & 54.8 & 52.32 & 19.95 \\
60\% & IQP-noHH & 59.4 & 30.6 & 47.0 & 69.9 & 53.3 & 52.03 & 20.96 \\
60\% & Add-count & 59.9 & 31.6 & 47.3 & 70.0 & 53.1 & 52.38 & 21.09 \\
60\% & Pair-count & 58.4 & 32.1 & 47.2 & 70.1 & 54.3 & 52.41 & 19.83 \\
60\% & ShortGPT & 50.6 & 27.7 & 42.7 & 65.9 & 52.6 & 47.91 & 26.85 \\
60\% & SLEB & 56.1 & 30.8 & 45.1 & 68.7 & 51.6 & 50.47 & 21.43 \\
\bottomrule
\end{tabular}
\end{table}
\begin{table}[H]\centering\caption{Qwen2.5-1.5B, LoRA.}\label{tab:pt15l}\footnotesize
\begin{tabular}{llccccccc}
\toprule
budget & structure & ARC-e & ARC-c & HellaSwag & PIQA & WinoG. & macro & ppl \\
\midrule
80\% & ILP & 69.2 & 38.7 & 58.2 & 74.2 & 56.2 & 59.29 & 15.47 \\
80\% & IQP & 66.4 & 39.5 & 58.8 & 73.5 & 58.4 & 59.34 & 15.70 \\
80\% & Hybrid & 69.9 & 39.2 & 59.9 & 74.3 & 59.1 & 60.47 & 15.62 \\
80\% & IQP-cap & 70.4 & 40.2 & 59.1 & 73.8 & 58.8 & 60.45 & 15.16 \\
80\% & IQP-noHH & 68.2 & 39.6 & 59.6 & 73.9 & 60.1 & 60.28 & 15.75 \\
80\% & Add-count & 65.1 & 37.2 & 58.0 & 73.4 & 57.2 & 58.20 & 15.09 \\
80\% & Pair-count & 68.3 & 38.3 & 58.7 & 74.2 & 57.6 & 59.40 & 14.54 \\
\midrule
70\% & ILP & 62.2 & 33.5 & 51.0 & 70.6 & 56.4 & 54.76 & 19.33 \\
70\% & IQP & 63.1 & 34.4 & 52.7 & 71.5 & 56.6 & 55.66 & 17.65 \\
70\% & Hybrid & 64.3 & 35.5 & 53.0 & 71.8 & 57.3 & 56.37 & 18.70 \\
70\% & IQP-cap & 65.2 & 37.0 & 53.4 & 72.2 & 54.5 & 56.47 & 17.92 \\
70\% & IQP-noHH & 63.8 & 34.2 & 52.6 & 71.2 & 57.5 & 55.86 & 18.72 \\
70\% & Add-count & 63.8 & 35.7 & 52.4 & 71.4 & 54.5 & 55.55 & 18.37 \\
70\% & Pair-count & 61.6 & 34.3 & 52.6 & 71.0 & 55.6 & 55.02 & 17.36 \\
\midrule
60\% & ILP & 55.2 & 30.2 & 44.0 & 68.4 & 54.0 & 50.36 & 24.60 \\
60\% & IQP & 56.6 & 31.0 & 46.2 & 70.2 & 53.9 & 51.58 & 21.07 \\
60\% & Hybrid & 55.5 & 29.7 & 44.4 & 69.1 & 53.7 & 50.46 & 24.86 \\
60\% & IQP-cap & 58.5 & 32.0 & 45.9 & 69.5 & 54.9 & 52.16 & 20.99 \\
60\% & IQP-noHH & 57.4 & 30.6 & 46.2 & 69.4 & 53.0 & 51.32 & 21.95 \\
60\% & Add-count & 57.9 & 30.6 & 46.1 & 69.4 & 53.6 & 51.51 & 22.11 \\
60\% & Pair-count & 56.6 & 31.3 & 46.3 & 69.7 & 52.8 & 51.35 & 20.57 \\
\bottomrule
\end{tabular}
\end{table}
\begin{table}[H]\centering\caption{Qwen2.5-3B, LoRA.}\label{tab:pt3b}\footnotesize
\begin{tabular}{llccccccc}
\toprule
budget & structure & ARC-e & ARC-c & HellaSwag & PIQA & WinoG. & macro & ppl \\
\midrule
80\% & ILP & 73.0 & 44.5 & 68.4 & 76.9 & 62.6 & 65.10 & 12.37 \\
80\% & IQP & 72.6 & 44.2 & 65.8 & 76.4 & 58.1 & 63.41 & 14.06 \\
80\% & Hybrid & 72.7 & 44.5 & 68.4 & 76.8 & 64.3 & 65.34 & 12.42 \\
80\% & IQP-cap & 73.3 & 44.2 & 66.4 & 76.8 & 60.9 & 64.31 & 12.57 \\
80\% & IQP-noHH & 74.2 & 45.5 & 66.8 & 75.5 & 64.4 & 65.28 & 13.43 \\
80\% & Add-count & 75.1 & 45.7 & 68.5 & 77.2 & 63.4 & 65.97 & 13.00 \\
80\% & Pair-count & 73.0 & 44.8 & 66.3 & 77.0 & 62.7 & 64.75 & 12.64 \\
80\% & ShortGPT & 66.3 & 38.9 & 61.7 & 75.5 & 59.1 & 60.28 & 14.63 \\
80\% & SLEB & 69.1 & 40.7 & 63.7 & 76.7 & 58.8 & 61.83 & 12.65 \\
\midrule
70\% & ILP & 69.9 & 41.0 & 62.1 & 75.0 & 61.5 & 61.90 & 14.93 \\
70\% & IQP & 65.0 & 38.4 & 60.2 & 74.8 & 56.1 & 58.90 & 15.81 \\
70\% & Hybrid & 69.7 & 41.4 & 63.1 & 74.7 & 61.0 & 61.98 & 14.27 \\
70\% & IQP-cap & 64.7 & 37.4 & 59.9 & 75.3 & 59.7 & 59.41 & 15.43 \\
70\% & IQP-noHH & 66.2 & 39.9 & 60.2 & 74.5 & 55.9 & 59.36 & 16.26 \\
70\% & Add-count & 68.8 & 39.3 & 60.9 & 75.2 & 58.2 & 60.47 & 15.33 \\
70\% & Pair-count & 66.4 & 38.7 & 59.7 & 74.0 & 57.6 & 59.28 & 15.40 \\
70\% & ShortGPT & 66.3 & 36.5 & 57.3 & 73.8 & 55.7 & 57.92 & 15.86 \\
70\% & SLEB & 67.3 & 36.9 & 56.6 & 74.6 & 53.9 & 57.87 & 14.84 \\
\midrule
60\% & ILP & 62.8 & 35.1 & 53.5 & 71.5 & 58.0 & 56.16 & 18.90 \\
60\% & IQP & 62.6 & 34.9 & 53.4 & 71.7 & 53.9 & 55.33 & 18.91 \\
60\% & Hybrid & 62.5 & 34.9 & 54.7 & 72.3 & 56.1 & 56.09 & 18.43 \\
60\% & IQP-cap & 60.6 & 33.5 & 52.2 & 72.3 & 55.5 & 54.82 & 18.61 \\
60\% & IQP-noHH & 59.9 & 33.2 & 52.9 & 71.7 & 54.4 & 54.42 & 18.05 \\
60\% & Add-count & 64.2 & 34.8 & 55.4 & 73.3 & 56.8 & 56.88 & 17.77 \\
60\% & Pair-count & 61.3 & 34.0 & 52.9 & 71.9 & 53.8 & 54.78 & 18.69 \\
60\% & ShortGPT & 56.0 & 30.6 & 45.5 & 67.5 & 51.7 & 50.26 & 24.16 \\
60\% & SLEB & 59.7 & 32.0 & 48.4 & 71.9 & 52.2 & 52.86 & 19.10 \\
\bottomrule
\end{tabular}
\end{table}

\section{Held-out task examples}
\label{sec:split}
Using the same task examples for the short-recovery signal and for the final accuracy overstates the signal
(Table~\ref{tab:split}). All selection results in the main text use disjoint halves.

\begin{table}[H]
\centering
\caption{Short-recovery zero-shot accuracy as a ranking signal: decided pairs ordered correctly when the signal and the
final accuracy use the same examples, and when they use disjoint halves (both directions pooled).}
\label{tab:split}
\footnotesize
\begin{tabular}{llcc}
\toprule
setting & steps & same examples & disjoint halves \\
\midrule
1.5B full & 100 & 33/33 = 100\% & 61/65 = 94\% \\
 & 400 & 33/33 = 100\% & 58/65 = 89\% \\
1.5B LoRA & 100 & 45/45 = 100\% & 75/88 = 85\% \\
 & 400 & 45/45 = 100\% & 76/88 = 86\% \\
3B LoRA & 100 & 46/48 = 96\% & 81/92 = 88\% \\
 & 400 & 48/48 = 100\% & 85/92 = 92\% \\
\bottomrule
\end{tabular}

\end{table}

\section{Stacking probe}
\label{sec:stackdetail}
For each block type, the 16 layers with the lowest single-block KL are replaced one at a time, in pairs (all pairs measured), in nested prefixes of three random orders and in 40 random
triples. For halved attention the median pair interaction is $-9.2\%$ (1.5B) and $-11.3\%$ (3B) of the additive
prediction and the median triple residual $+7.7\%$ and $+8.8\%$. For the 23 halved-attention layers chosen by the IQP on
the 3B model at 80\%, the measured KL is 0.121, the additive prediction 0.102, the pairwise prediction from the
measured pair table (27\% of the pairs) 0.057 and from all pairs $-0.083$.

\begin{figure}[H]
\centering
\includegraphics[width=\linewidth]{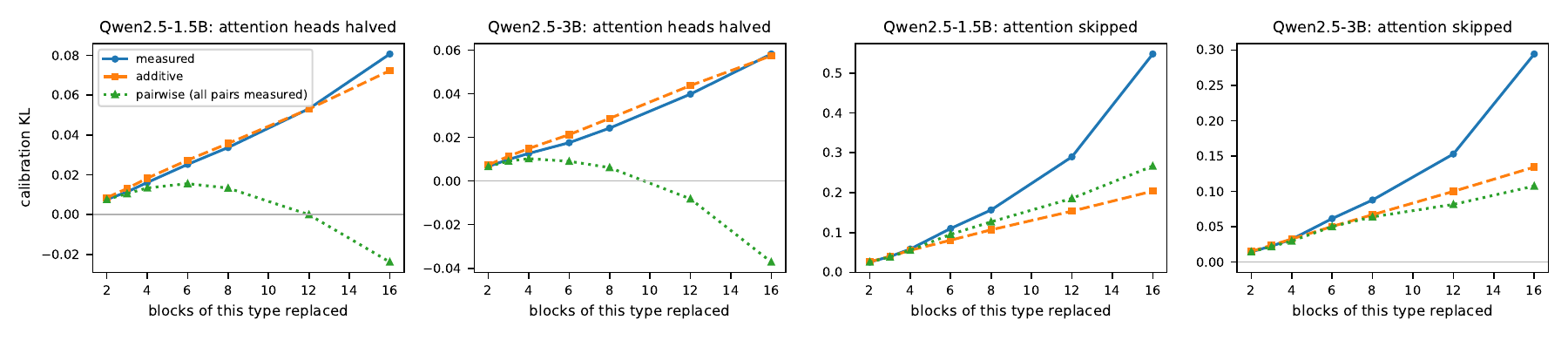}
\caption{Measured calibration KL when one block type replaces a growing set of layers (mean of three random orders),
against the additive and the pairwise prediction with every pair measured.}
\label{fig:stack}
\end{figure}

\begin{table}[H]
\centering
\caption{Sixteen layers of one block type (the 16 with the lowest single-block KL): measured KL and the relative error
of the additive and pairwise predictions.}
\label{tab:stack}
\footnotesize
\begin{tabular}{lcccccc}
\toprule
 & \multicolumn{3}{c}{Qwen2.5-1.5B} & \multicolumn{3}{c}{Qwen2.5-3B} \\
\cmidrule(lr){2-4}\cmidrule(lr){5-7}
block type & real KL & additive & pairwise & real KL & additive & pairwise \\
\midrule
attention heads halved & 0.081 & -10\% & -130\% & 0.058 & -1\% & -164\% \\
attention skipped & 0.548 & -63\% & -51\% & 0.294 & -54\% & -63\% \\
MLP pruned 50\% & 0.327 & -22\% & -24\% & 0.203 & -17\% & -52\% \\
MLP pruned 75\% & 0.707 & -31\% & +3\% & 0.450 & -34\% & -20\% \\
MLP low-rank & 0.331 & -7\% & -19\% & 0.186 & +5\% & -50\% \\
MLP skipped & 1.254 & -26\% & +0\% & 1.313 & -59\% & -45\% \\
block skipped & 2.592 & -37\% & +36\% & 1.707 & -41\% & -3\% \\
\bottomrule
\end{tabular}

\end{table}

\begin{table}[H]
\centering
\caption{Sixteen layers of one block type, before recovery: calibration KL (measured / additive prediction) and drop of
zero-shot accuracy (points, first 500 examples of every task) measured and predicted as the sum of single-layer drops.}
\label{tab:stacktask}
\footnotesize\setlength{\tabcolsep}{3pt}
\begin{tabular}{lcccccc}
\toprule
 & \multicolumn{3}{c}{Qwen2.5-1.5B} & \multicolumn{3}{c}{Qwen2.5-3B} \\
\cmidrule(lr){2-4}\cmidrule(lr){5-7}
block type & KL real / add. & acc.\ drop real & acc.\ drop add. & KL real / add. & acc.\ drop real & acc.\ drop add. \\
\midrule
attention heads halved & 0.081 / 0.072 & 1.56 & 0.56 & 0.058 / 0.057 & 0.52 & 5.64 \\
attention skipped & 0.548 / 0.204 & 12.44 & 7.20 & 0.294 / 0.134 & 3.96 & 3.16 \\
MLP pruned 50\% & 0.327 / 0.255 & 6.92 & 1.16 & 0.203 / 0.169 & 4.16 & 6.60 \\
\bottomrule
\end{tabular}

\end{table}

\section{Pre-specification of the stability diagnostic}
\label{sec:prereg}
The diagnostic is $D=\mathrm{Spearman}(\KL_0,\KL_{25})$ over the seven searched structures of a cell (ILP, IQP, hybrid,
IQP-cap, IQP-noHH, Add-count, Pair-count), where $\KL_s$ is the calibration KL after $s$ steps of the proxy recovery
(seed-averaged). Fixed on 27 November 2026 at 12:40 JST (file time stamp; not publicly deposited), before any step-25 data of a new model existed: threshold
$t=0.875$; rule as in \cref{sec:analysis}; prediction 3a, all three Llama-3.2-1B cells have $D\ge t$ (3/3 pass, 2/3
partial, at most 1/3 fail and the diagnostic is dropped as a method); prediction 3b, on models not yet run, the pooled
pre-recovery zero-shot pair accuracy (held-out halves) is at least 75\% in stable cells and at most 60\% in unstable
cells. The specification, the analysis script and the result files are archived with SHA-256 hashes (specification
\texttt{17fae153\ldots}, analysis script \texttt{9e80b236\ldots}; full hashes and files available from the authors).
Results: prediction 3a, two of three cells (partial pass; \cref{sec:analysis}); prediction 3b on Llama-3.2-3B, all
cells stable and 81\% in stable cells, unstable side untested.

\section{Surrogates for large structures}
\label{sec:surrogate}
Prefixes (25, 50, 75 and 100\% of the replaced blocks, two random orders) of every searched structure are measured,
giving structures with up to 35 replaced blocks; the score tables contain combinations of at most six blocks. Table
\ref{tab:surr} fits each surrogate on the combinations and on the prefixes of the structures of two budgets and tests
it on the full structures of the third.

\begin{table}[H]
\centering
\caption{Surrogates of the calibration KL, leave-one-budget-out: median relative error on all test structures and on
those with at least 15 replaced blocks, and mean within-budget Spearman correlation with the measured KL.}
\label{tab:surr}
\footnotesize
\begin{tabular}{lcccccc}
\toprule
 & \multicolumn{3}{c}{Qwen2.5-1.5B} & \multicolumn{3}{c}{Qwen2.5-3B} \\
\cmidrule(lr){2-4}\cmidrule(lr){5-7}
surrogate & rel.\ error & $k\ge15$ & Spearman & rel.\ error & $k\ge15$ & Spearman \\
\midrule
additive & 19.3\% & 21.4\% & 0.67 & 19.6\% & 24.1\% & 0.61 \\
pairwise & 21.0\% & 29.2\% & 0.71 & 13.5\% & 36.9\% & 0.68 \\
scaled additive & 18.1\% & 18.7\% & 0.67 & 21.2\% & 21.8\% & 0.61 \\
pairwise, $1/(k-1)$ shrinkage & 18.3\% & 23.1\% & 0.67 & 20.4\% & 25.0\% & 0.63 \\
additive + quadratic & 19.1\% & 21.3\% & 0.67 & 19.3\% & 23.5\% & 0.61 \\
additive + per-type count & 11.9\% & 9.8\% & 0.79 & 12.3\% & 11.5\% & 0.67 \\
pairwise + per-type count & 6.9\% & 9.4\% & 0.77 & 11.1\% & 14.2\% & 0.84 \\
\bottomrule
\end{tabular}

\end{table}

\section{Count-corrected and constrained searches on Qwen2.5-3B}
\label{sec:countdetail}
\begin{table}[H]
\centering
\caption{Qwen2.5-3B: differences in final accuracy (points) with paired bootstrap 95\% intervals over examples
(seed-averaged; conditional on the two recovery seeds). $^*$: interval excludes zero.}
\label{tab:count}
\footnotesize
\textbf{(a) count-corrected surrogates}\\[2pt]
\begin{tabular}{llccc}
\toprule
budget & structure & $-$ ILP & $-$ IQP & $-$ Hybrid \\
\midrule
80\% & Add-count & +0.87 [+0.30, +1.43]$^*$ & +2.56 [+1.86, +3.24]$^*$ & +0.63 [+0.001, +1.26]$^*$ \\
80\% & Pair-count & -0.35 [-1.03, +0.33] & +1.34 [+0.62, +2.06]$^*$ & -0.59 [-1.25, +0.09] \\
70\% & Add-count & -1.43 [-2.10, -0.72]$^*$ & +1.57 [+0.90, +2.23]$^*$ & -1.51 [-2.22, -0.79]$^*$ \\
70\% & Pair-count & -2.62 [-3.37, -1.88]$^*$ & +0.38 [-0.31, +1.06] & -2.70 [-3.43, -1.97]$^*$ \\
60\% & Add-count & +0.73 [-0.01, +1.45] & +1.56 [+0.82, +2.31]$^*$ & +0.79 [+0.001, +1.60]$^*$ \\
60\% & Pair-count & -1.38 [-2.11, -0.65]$^*$ & -0.55 [-1.26, +0.17] & -1.31 [-2.08, -0.54]$^*$ \\
\bottomrule
\end{tabular}
\\[6pt]
\textbf{(b) IQP with halved attention capped at the ILP's count (IQP-cap) or without halved-attention pair terms (IQP-noHH)}\\[2pt]
\begin{tabular}{llccc}
\toprule
budget & structure & $-$ ILP & $-$ IQP & $-$ Hybrid \\
\midrule
80\% & IQP-cap & -0.79 [-1.43, -0.15]$^*$ & +0.90 [+0.15, +1.65]$^*$ & -1.04 [-1.72, -0.36]$^*$ \\
80\% & IQP-noHH & +0.17 [-0.50, +0.87] & +1.87 [+1.13, +2.59]$^*$ & -0.07 [-0.73, +0.60] \\
70\% & IQP-cap & -2.49 [-3.26, -1.73]$^*$ & +0.51 [-0.24, +1.26] & -2.57 [-3.33, -1.80]$^*$ \\
70\% & IQP-noHH & -2.54 [-3.30, -1.77]$^*$ & +0.46 [-0.24, +1.16] & -2.62 [-3.39, -1.83]$^*$ \\
60\% & IQP-cap & -1.33 [-2.04, -0.61]$^*$ & -0.50 [-1.24, +0.25] & -1.26 [-2.06, -0.44]$^*$ \\
60\% & IQP-noHH & -1.74 [-2.54, -0.92]$^*$ & -0.91 [-1.66, -0.15]$^*$ & -1.67 [-2.46, -0.87]$^*$ \\
\bottomrule
\end{tabular}

\end{table}

\section{Concentration of the divergence: robustness}
\label{sec:conc}
The concentration statistic of Section~\ref{sec:analysis} was found after the fact among 14 statistics of the decomposition. To check
that it does not depend on the threshold, for each structure included in the recovery experiments the token-level KL was computed on the
calibration set using its pre-recovery checkpoint (the library replacements before any recovery training) and summarised by a family of statistics fixed before looking at the result, with the top-10\% share
as the primary one (Table~\ref{tab:conc}). All statistics use calibration data only, so the final accuracy is
evaluated on all task examples. Every concentration statistic orders the decided pairs significantly better than chance
on all three settings, except the top-1\% share on the 3B model ($p=0.06$), while the mean of the same token-level KL,
the search objective, does not. The concentration is nearly uncorrelated with the mean KL (within-budget rank
correlation $\le0.07$) and with the number of replaced blocks; controlling for both leaves a partial rank correlation
with the final accuracy of 0.56 (1.5B full), 0.72 (1.5B LoRA) and 0.66 (3B). Leaving out any one search method keeps
73--96\% of the decided pairs ordered correctly. On the 1.5B model the concentration tracks the pre-recovery accuracy
(within-budget rank correlation 0.55), on the 3B model it does not ($-0.10$) but tracks the gain during recovery
($+0.30$). As a selection rule in the prospective run it is competitive but not better than the other rules
(\cref{tab:prospective}). With the criterion fixed in advance it replicated on Llama-3.2-1B and failed on
Llama-3.2-3B (\cref{sec:analysis}). Table~\ref{tab:conc} uses calibration data only and the final accuracy on all
examples; Table~\ref{tab:decomp} scores the final accuracy on the examples the decomposition does not use, so the two
tables differ slightly for the same statistic.

\begin{table}[H]
\centering
\caption{Statistics of the pre-recovery token-level KL as ranking signals among the seven searched structures of a
budget: decided pairs ordered correctly (permutation $p$). Higher concentration ranks higher; lower mean KL ranks
higher.}
\label{tab:conc}
\footnotesize
\begin{tabular}{lccc}
\toprule
statistic of the token-level KL & 1.5B full & 1.5B LoRA & 3B LoRA \\
\midrule
top-10\% share (primary) & 85\% (0.005) & 82\% (0.003) & 79\% (0.005) \\
top-1\% share & 94\% (0.0002) & 87\% (0.0008) & 69\% (0.056) \\
top-5\% share & 88\% (0.002) & 84\% (0.003) & 79\% (0.004) \\
top-20\% share & 85\% (0.006) & 82\% (0.004) & 81\% (0.001) \\
Gini coefficient & 85\% (0.004) & 80\% (0.005) & 83\% (0.001) \\
1 $-$ normalised entropy & 85\% (0.003) & 80\% (0.005) & 83\% (0.001) \\
mean KL (search objective) & 67\% (0.125) & 58\% (0.280) & 31\% (0.961) \\
\bottomrule
\end{tabular}

\end{table}

\section{Recovery gains differ by structure}
\label{sec:gains}
On the 3B model the ILP structure at 70\% gains 14.3 points during recovery and the hybrid's 4.3, ending level at 61.9
and 62.0 although they started 10 points apart.

\section{Common-protocol sensitivity and paired comparisons}
\label{sec:sens}
Table~\ref{tab:sens} repeats the common protocol of \cref{tab:families} with every pair counted (threshold 0), with a
1-point threshold, and leaving one task out: task-based signals are computed on four tasks (all examples), the final
accuracy on the fifth, pooled over the five choices of held-out task with one shuffle per budget shared by all folds.
Table~\ref{tab:paired} compares zero-shot accuracy after 100 steps with each other signal on the same decided pairs
(0.5-point threshold, held-out halves): the difference in the share of correctly ordered pairs, with a sign-flip
randomisation test that exchanges the two signals within each budget and split direction (six units per setting,
exact; 36 units pooled, 20{,}000 draws). Seed standard deviations of the final accuracy (mean over structures):
Qwen2.5-0.5B 0.25, 1.5B full 0.28, 1.5B LoRA 0.19, 3B 0.18, Llama-3.2-1B 0.19 (three seeds), 3B 0.17 points.

\begin{table}[H]
\centering
\caption{Common protocol, sensitivity: decided pairs ordered correctly (\%, permutation $p$). Bold: $p<0.05$.}
\label{tab:sens}
\footnotesize\setlength{\tabcolsep}{3pt}
\resizebox{\textwidth}{!}{\begin{tabular}{lcccccc}
\toprule
Signal, step & Qwen 0.5B full & Qwen 1.5B full & Qwen 1.5B LoRA & Qwen 3B LoRA & Llama 1B LoRA & Llama 3B LoRA \\
\midrule
\multicolumn{7}{l}{\textit{(a) every pair (threshold 0)}} \\
KL, before & \textbf{69 (.043)} & 64 (.063) & 57 (.22) & 34 (.97) & 49 (.59) & 64 (.067) \\
Zero-shot, before & \textbf{67 (.042)} & \textbf{76 ($<$.001)} & \textbf{75 ($<$.001)} & 36 (.96) & \textbf{68 (.014)} & \textbf{75 (.002)} \\
KL, 100 & \textbf{74 (.008)} & 64 (.065) & 62 (.092) & \textbf{68 (.026)} & 60 (.17) & \textbf{68 (.029)} \\
Loss, 100 & \textbf{74 (.008)} & 62 (.092) & 59 (.17) & \textbf{66 (.038)} & 55 (.32) & \textbf{66 (.045)} \\
Loss, 400 & \textbf{74 (.008)} & 64 (.062) & 60 (.13) & \textbf{71 (.010)} & 60 (.17) & \textbf{69 (.018)} \\
Zero-shot, 100 & 61 (.13) & \textbf{82 ($<$.001)} & \textbf{79 ($<$.001)} & \textbf{83 ($<$.001)} & \textbf{71 (.010)} & \textbf{75 (.002)} \\
Zero-shot, 400 & 63 (.091) & \textbf{75 (.001)} & \textbf{79 ($<$.001)} & \textbf{85 ($<$.001)} & \textbf{71 (.008)} & \textbf{75 ($<$.001)} \\
Decided pairs & 70 & 126 & 126 & 126 & 92 & 114 \\
\midrule
\multicolumn{7}{l}{\textit{(b) final gap $>1$ point}} \\
KL, before & 88 (.051) & 58 (.34) & 63 (.19) & 35 (.88) & 60 (.39) & \textbf{80 (.021)} \\
Zero-shot, before & \textbf{88 (.039)} & \textbf{87 (.011)} & \textbf{84 (.003)} & 32 (.93) & 87 (.065) & \textbf{96 ($<$.001)} \\
KL, 100 & 88 (.051) & 53 (.46) & 69 (.096) & \textbf{82 (.005)} & 73 (.22) & \textbf{85 (.009)} \\
Loss, 100 & 88 (.051) & 50 (.53) & 61 (.23) & \textbf{82 (.006)} & 73 (.22) & \textbf{80 (.022)} \\
Loss, 400 & \textbf{94 (.023)} & 58 (.34) & 67 (.12) & \textbf{88 ($<$.001)} & 87 (.10) & \textbf{85 (.008)} \\
Zero-shot, 100 & \textbf{88 (.036)} & \textbf{97 ($<$.001)} & \textbf{96 ($<$.001)} & \textbf{92 ($<$.001)} & \textbf{100 (.011)} & \textbf{91 (.001)} \\
Zero-shot, 400 & 75 (.11) & \textbf{97 ($<$.001)} & \textbf{94 ($<$.001)} & \textbf{96 ($<$.001)} & \textbf{100 (.009)} & \textbf{94 ($<$.001)} \\
Decided pairs & 16 & 38 & 51 & 66 & 15 & 46 \\
\midrule
\multicolumn{7}{l}{\textit{(c) leave one task out (gap $>0.5$)}} \\
KL, before & 65 (.050) & 59 (.13) & 57 (.20) & 35 (.96) & 47 (.63) & 60 (.11) \\
Zero-shot, before & \textbf{68 (.024)} & \textbf{72 ($<$.001)} & \textbf{74 (.001)} & 38 (.93) & \textbf{73 (.014)} & \textbf{73 ($<$.001)} \\
KL, 100 & \textbf{68 (.021)} & 60 (.10) & 62 (.073) & \textbf{69 (.009)} & 61 (.17) & \textbf{67 (.022)} \\
Loss, 100 & \textbf{68 (.021)} & 58 (.14) & 58 (.17) & \textbf{68 (.014)} & 58 (.26) & \textbf{65 (.034)} \\
Loss, 400 & \textbf{68 (.021)} & \textbf{63 (.046)} & 60 (.12) & \textbf{74 (.001)} & 63 (.14) & \textbf{69 (.011)} \\
Zero-shot, 100 & \textbf{68 (.024)} & \textbf{75 ($<$.001)} & \textbf{79 ($<$.001)} & \textbf{78 ($<$.001)} & \textbf{72 (.021)} & \textbf{70 (.006)} \\
Zero-shot, 400 & 62 (.077) & \textbf{69 (.004)} & \textbf{79 ($<$.001)} & \textbf{79 ($<$.001)} & \textbf{72 (.017)} & \textbf{69 (.006)} \\
Decided pairs & 112 & 206 & 223 & 232 & 131 & 210 \\
\bottomrule
\end{tabular}
}
\end{table}

\begin{table}[H]
\centering
\caption{Zero-shot accuracy after 100 steps minus each other signal: difference in decided pairs ordered correctly
(percentage points, same pairs) and sign-flip $p$ (two-sided). With six units the smallest attainable $p$ is 0.031.
Bold: $p<0.05$.}
\label{tab:paired}
\footnotesize\setlength{\tabcolsep}{3pt}
\resizebox{\textwidth}{!}{\begin{tabular}{lccccccc}
\toprule
Zero-shot, 100 minus & Qwen 0.5B full & Qwen 1.5B full & Qwen 1.5B LoRA & Qwen 3B LoRA & Llama 1B LoRA & Llama 3B LoRA & pooled \\
\midrule
KL, before & +10 (.50) & +31 (.25) & \textbf{+24 (.031)} & \textbf{+56 (.031)} & +34 (.25) & +12 (.062) & \textbf{+30 ($<$.001)} \\
Zero-shot, before & +3 (1.00) & +6 (.50) & +3 (.69) & \textbf{+56 (.031)} & +7 (.75) & -5 (.88) & \textbf{+14 (.010)} \\
KL, 100 & +8 (.62) & +32 (.25) & +16 (.062) & +12 (.12) & +18 (.25) & +9 (.12) & \textbf{+16 ($<$.001)} \\
Loss, 100 & +8 (.62) & +35 (.25) & +20 (.062) & +14 (.12) & +25 (.19) & +10 (.12) & \textbf{+19 ($<$.001)} \\
KL, 400 & +5 (.88) & +29 (.25) & +17 (.062) & +4 (.50) & +18 (.25) & +5 (.50) & \textbf{+13 ($<$.001)} \\
Loss, 400 & +5 (.88) & +26 (.25) & \textbf{+19 (.031)} & +5 (.50) & +18 (.25) & +6 (.50) & \textbf{+13 ($<$.001)} \\
Zero-shot, 400 & +15 (.34) & +5 (.25) & -1 (1.00) & -4 (.50) & +4 (.75) & -4 (.62) & +1 (.84) \\
\bottomrule
\end{tabular}
}
\end{table}

\section{Prospective pools with the searched structures added}
\label{sec:aug}
Adding the ILP, IQP and hybrid structures (with their existing 400-step proxy runs read at step 100) to the 16-member
pools changes the candidates but not the recipe. On Qwen2.5-3B the zero-shot rule then picks a pool member at 80\% (+0.28
points against the hybrid, not significant), the hybrid at 70\% and the pool's surrogate optimum at 60\% (+1.39 against
the hybrid, interval excludes zero); picking by the pre-recovery KL is 1.01 points below the hybrid at 80\% (interval
excludes zero). The 60\% gain comes from the pool construction, since the surrogate optimum is picked without any proxy
recovery. On Llama-3.2-1B (three seeds) the zero-shot rule picks the hybrid, the IQP and a pool member and is never
significantly different from the ILP, the IQP or the hybrid; at 70\% four of the six rules pick the IQP, which is 1.05
points above the hybrid (interval excludes zero). On Qwen2.5-0.5B the 100-step rule then picks the ILP, the hybrid and
the IQP and is never significantly below another searched structure. The zero-step rule, with the pre-recovery
accuracy of the searched structures on the same even half, picks the ILP, the IQP and a pool member on Llama-3.2-1B
(never significantly below a searched structure) and the hybrid, the IQP and a pool member on Qwen2.5-0.5B, where the
IQP at 70\% is 1.40 points below the hybrid (interval excludes zero).

\begin{table}[h]
\centering
\caption{Prospective two-stage selection on Llama-3.2-1B (LoRA, three seeds) and Qwen2.5-0.5B (full fine-tuning, two
seeds): final accuracy (\%, odd-indexed half) of the searched structures and of the pool member picked by each rule
(16-member pools). Bold: best of the budget. $^-$: significantly below at least one searched structure (paired
bootstrap 95\% interval excludes zero); no pick is significantly above all three. The pre-recovery zero-shot rule is
the zero-step rule of the prospective test in \cref{sec:select}.}
\label{tab:prosp2}
\scriptsize\setlength{\tabcolsep}{2pt}
\resizebox{\textwidth}{!}{\begin{tabular}{lcccccccc}
\toprule
 & \multicolumn{4}{c}{Llama-3.2-1B (LoRA, 3 seeds)} & \multicolumn{4}{c}{Qwen2.5-0.5B (full FT, 2 seeds)} \\
\cmidrule(lr){2-5}\cmidrule(lr){6-9}
structure & 80\% & 70\% & 60\% & mean & 80\% & 70\% & 60\% & mean \\
\midrule
searched: ILP & 54.61 & 51.10 & 45.71 & 50.47 & 51.43 & 49.13 & 45.33 & 48.63 \\
searched: IQP & \textbf{54.98} & \textbf{51.80} & \textbf{46.93} & \textbf{51.24} & \textbf{51.68} & 48.30 & 46.55 & 48.84 \\
searched: Hybrid & 54.61 & 50.75 & 45.88 & 50.41 & 51.44 & \textbf{49.70} & 45.76 & 48.97 \\
\midrule
pool: surrogate optimum & 54.55 & 50.00$^-$ & 46.37 & 50.31 & 50.84 & 47.85$^-$ & 46.88 & 48.52 \\
pool: pre-recovery calib.\ KL (min) & 54.55 & 50.00$^-$ & 46.85 & 50.47 & 51.50 & 49.29 & 46.98 & \textbf{49.26} \\
pool: pre-recovery zero-shot acc.\ (even half) & 54.91 & 51.01 & 46.58 & 50.83 & 50.96 & 48.83 & 47.20 & 49.00 \\
pool: concentration (max) & 54.91 & 50.48$^-$ & 46.65 & 50.68 & 50.70$^-$ & 49.29 & 46.88 & 48.96 \\
pool: KL after 100 steps (min) & 54.55 & 50.62$^-$ & 46.85 & 50.67 & 50.96 & 48.03$^-$ & 46.98 & 48.66 \\
pool: KD loss after 100 steps (min) & 54.55 & 50.62$^-$ & 46.85 & 50.67 & 50.96 & 48.03$^-$ & 46.98 & 48.66 \\
pool: zero-shot acc.\ after 100 steps (even half) & 54.65 & 51.01 & 46.06 & 50.57 & 50.60$^-$ & 47.85$^-$ & \textbf{47.29} & 48.58 \\
\bottomrule
\end{tabular}
}
\end{table}

\section{Proxy-run divergence}
\label{sec:diverge}
The proxy run of the IQP structure at 80\% on the 3B model reached a training loss of 4.6 between steps 200 and 400
(zero-shot accuracy 37.8 at step 400); the corresponding recovery run did not diverge. Two reruns (seeds 0 and 1, fresh
LoRA initialisation) reached 62.9 and 62.5 at step 400 against a final 63.4. The structure is last in its budget under
every version, so all statistics are unchanged.

\end{document}